\documentclass[10pt,twocolumn,letterpaper]{article}

\usepackage[pagenumbers]{iccv} 
\usepackage{amsmath}
\usepackage{graphicx}
\usepackage{dsfont}
\usepackage{wrapfig}
\usepackage{multirow}
\usepackage{arydshln}
\usepackage{utfsym}
\usepackage{multirow}
\usepackage{array}
\usepackage{colortbl}
\usepackage{csquotes}
\usepackage{float}
\usepackage{placeins}
\newcommand{\vpara}[1]{\vspace{0.05in}\noindent\textbf{#1 }}

\definecolor{iccvblue}{rgb}{0.21,0.49,0.74}
\usepackage[pagebackref,breaklinks,colorlinks,allcolors=iccvblue]{hyperref}
\newcommand{\name}{LVBench}
\def\paperID{*****} 
\def\confName{ICCV}
\def\confYear{2025}

\title{{\name}: An Extreme \underline{L}ong \underline{V}ideo Understanding \underline{Bench}mark}

\author{Weihan Wang\textsuperscript{1} \ \ Zehai He\textsuperscript{2} \ \ Wenyi Hong\textsuperscript{2} \ \ Yean Cheng\textsuperscript{1}\ \ Xiaohan Zhang\textsuperscript{1} \ \ \\ Ji Qi\textsuperscript{1} \ \  Ming Ding\textsuperscript{1} \ \ 
Xiaotao Gu\textsuperscript{1}\thanks{Corresponding authors}  \ \ Shiyu Huang\textsuperscript{1}\footnotemark[1] \ \ Bin Xu\textsuperscript{2} \ \ Yuxiao Dong\textsuperscript{2}\footnotemark[1] \ \ Jie Tang\textsuperscript{2}
\\
  \textsuperscript{1}Zhipu AI\ \ \textsuperscript{2}Tsinghua University\ \ 
  \\
  \texttt{weihan.wang@aminer.cn, xiaotao.gu@aminer.cn,}  \\ \texttt{shiyu.huang@zhipuai.cn, yuxiaod@tsinghua.edu.cn} \\
}

\begin{document}
\maketitle

\begin{abstract}
Recent progress in multimodal large language models has markedly enhanced the understanding of short videos (typically under one minute), and several evaluation datasets have emerged accordingly. However, these advancements fall short of meeting the demands of real-world applications such as embodied intelligence for long-term decision-making, in-depth movie reviews and discussions, and live sports commentary, all of which require comprehension of long videos spanning several hours. To address this gap, we introduce {\name}, a benchmark specifically designed for long video understanding. Our dataset comprises publicly sourced videos and encompasses a diverse set of tasks aimed at long video comprehension and information extraction. 
{\name} is designed to challenge multimodal models to demonstrate long-term memory and extended comprehension capabilities. Our extensive evaluations reveal that current multimodal models still underperform on these demanding long video understanding tasks. Through {\name}, we aim to spur the development of more advanced models capable of tackling the complexities of long video comprehension.
\end{abstract}    
\section{Introduction}
\begin{figure*}[h]
  \centering
    \includegraphics[width=\linewidth]{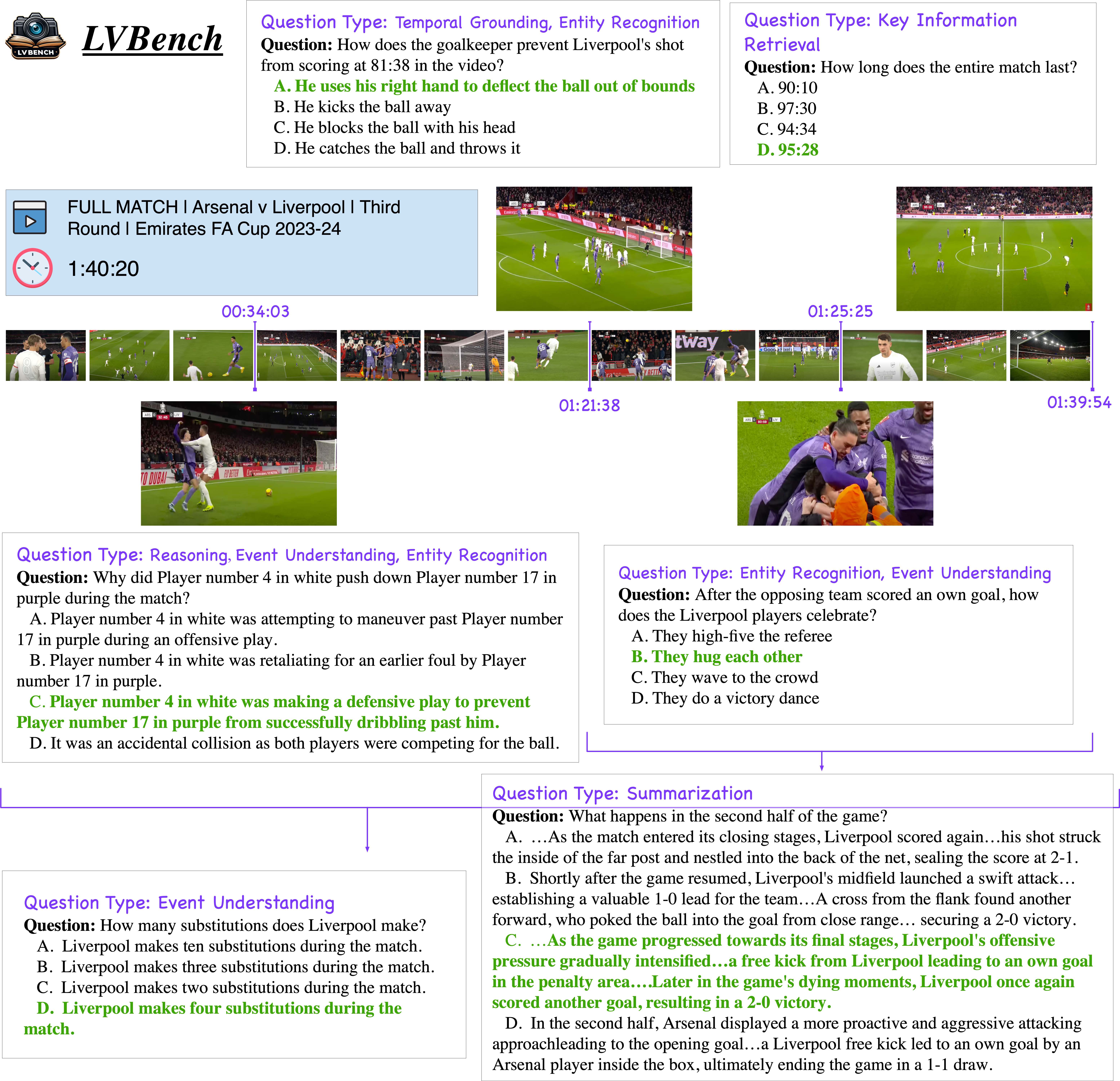} 
    \caption{Examples from {\name}. {\name} covers problems involving various temporal and spatial dimensions. 
    }
    \label{fig:examples}
\end{figure*}

The rapid development of Multi-modal Large Language Models (MLLMs) has led to significant advances in vision-language understanding~\citep{gpt4o, wang2024qwen2, wang2023cogvlm}, driven by breakthroughs in both large language models~\citep{openai2023gpt4,claude3,du2021glm} and visual encoders~\citep{radford2021learning,sun2023eva,zhai2023sigmoid}. These models have demonstrated impressive capabilities across various video understanding tasks, from fundamental tasks like action recognition and object tracking, to more complex tasks such as temporal event localization, dense video captioning, and abstractive video summarization. However, while existing MLLMs excel at processing short video clips, they face substantial challenges when dealing with longer temporal sequences - a crucial limitation that hinders their practical applications.
Despite numerous video understanding benchmarks being proposed, most primarily focus on evaluating spatial understanding in static images or short video clips, overlooking the critical aspect of temporal understanding in long-form videos. This gap is particularly significant given that real-world applications often require comprehension of extended temporal contexts. The scarcity of long video understanding benchmarks can be attributed to the challenges in data collection and the complexity of annotation.
To address these limitations, we introduce {\name}, a comprehensive benchmark designed specifically for evaluating MLLMs' capabilities in long video understanding. Our benchmark distinguishes itself from existing datasets through several key features:
\begin{itemize}
\item We establish a systematic framework of six core temporal understanding capabilities, which can be flexibly combined to create complex, multi-faceted evaluation tasks. This design enables a thorough assessment of models' ability to process and reason about extended temporal sequences.
\item We curate a diverse collection of long-form videos from various domains, with an average duration approximately four times longer than existing benchmarks. The videos span multiple categories (as illustrated in Figure~\ref{fig:examples}), providing a robust foundation for evaluating temporal understanding across different contexts.
\item Through a rigorous combination of expert human annotation and quality control processes, we ensure high-quality ground truth annotations, establishing a reliable benchmark for assessing long video understanding capabilities.
\end{itemize}

\section{Related Works}
\begin{table*}[t]
    \caption{Comparison of different datasets. \textbf{Open-domain} represents whether the video source is diversified. \textbf{Multi-type} represents whether the types of questions are greater than 2 categories.}
\begin{center}
        \begin{tabular}{ccccccc}
            \toprule
             \bf Dataset   & \bf Num QA & \bf Avg sec & \bf Open-domain & \textbf{Multi-type} & \bf Annotation \\
            \midrule
            TGIF-QA~\citep{jang2017tgif} & 165,165 & 3 & $\usym{2713}$ & $\usym{2717}$ & Auto \\
            MSVD-QA~\citep{xu2017video} & 13,157 & 10 & $\usym{2713}$ & $\usym{2717}$ & Auto \\
            MSRVTT-QA~\citep{xu2017video} & 72,821 & 15 & $\usym{2713}$ & $\usym{2717}$ & Auto\\
            MVBench~\citep{li2023mvbench}  &  4,000 & 16 & $\usym{2713}$ & $\usym{2713}$ & Auto  \\
            Perception Test~\citep{patraucean2023perception} &  44,000 & 23 & \usym{2717} & \usym{2713} &  Auto\&Manual \\
            NExT-QA~\citep{xiao2021next} & 52,044 & 44 & $\usym{2713}$ & $\usym{2717}$ & Manual \\
            How2QA~\citep{li2020hero} & 44,007 & 60 & $\usym{2713}$ & $\usym{2713}$ & Manual \\
            ActivityNet-QA~\citep{yu2019activitynet} & 800 & 111 & \usym{2717} & \usym{2717} & Manual \\
            CinePile~\citep{rawal2024cinepile}  &  {\bf303,828} & 160 & $\usym{2717}$ & $\usym{2713}$ & Auto\&Manual  \\
            EgoSchema~\citep{mangalam2024egoschema} &   5,000 &  180 & $\usym{2717}$ & $\usym{2713}$ & Auto\&Manual \\
            MovieQA~\citep{tapaswi2016movieqa} & 6,462 & 203 & $\usym{2717}$ & $\usym{2713}$ & Manual \\
            LongVideoBench~\citep{wu2024longvideobench} & 6,678 & 473 & $\usym{2713}$ & $\usym{2713}$ & Manual \\ 
            MovieChat-1K~\citep{song2023moviechat} & 13,000 & 564 & $\usym{2717}$ & $\usym{2713}$ & Manual \\
            MoVQA~\citep{zhang2023movqa}  &  21,953 & 992 & $\usym{2717}$ & $\usym{2713}$ & Manual  \\
            Video-MME~\citep{fu2024video} & 2,700 & 1018 & $\usym{2713}$ & $\usym{2713}$ & Manual \\  
            \midrule
            {\bf {\name}(Ours)} & 1,549 & {\bf 4,101}  & $\usym{2713}$ & $\usym{2713}$ & Manual \\    
            \bottomrule
        \end{tabular}
    \end{center}
    \label{table:dataset_compare}
\end{table*}
\subsection{Multi-modal Large Language Models}
Building upon the achievements in LLMs, the field has shifted towards MLLMs to enhance multi-modal understanding capabilities~\citep{alayrac2022flamingo,hong2023cogagent,wang2023cogvlm,gpt4o,reid2024gemini}. Early advancements in this area include models like Flamingo~\citep{alayrac2022flamingo}, which introduced a novel approach by adding cross-attention layers to a frozen language model, enabling efficient vision-language alignment without full model retraining. Recent works have made significant progress in video understanding through various architectural innovations. Video-LLaMA~\citep{zhang2023video} employs a video transformer with Q-Former for effective temporal modeling. Models like InternVL2~\citep{chen2024far}, LLaVA-Next~\citep{zhang2024llavanextvideo}, and CogVLM2~\citep{hong2024cogvlm2} unify high-resolution images and videos as multi-frame sequences, enabling seamless processing of both modalities. Qwen2-VL~\citep{wang2024qwen2} introduces 3D-RoPE to achieve strong temporal extrapolation capabilities for long video understanding. To handle extended temporal contexts, MovieChat~\citep{song2023moviechat} proposes a dual-memory architecture with short-term and long-term memory modules that can efficiently process videos over 10,000 frames. LWM\citep{liu2024world} with ring attention has pushed context lengths to the million-token scale, providing a promising foundation for long video understanding. PLLaVA\citep{xu2024pllava} introduced a resource-efficient method for adapting image-language pre-trained models to dense video understanding through a novel feature pooling strategy. 
Despite these significant architectural advances, our experiments indicate that current video understanding models still fall short on tasks requiring long-range comprehension, highlighting an urgent need for the development of models specifically tailored for long video understanding.

\subsection{ Benchmarks for MLLM}
{\bf} Recent advancements in vision-language benchmarks have largely focused on images and short videos, as seen in datasets like MMBench~\citep{liu2023mmbench}, SEED-Bench-2~\citep{li2023seedbench2}, TGIF-QA~\citep{jang2017tgif} and MVBench~\citep{li2023mvbench}. For long video understanding, previous benchmarks such as Perception Test~\citep{patraucean2023perception} have explored multi-modal video perception and reasoning but often with shorter video clips and limited temporal context. Datasets like How2QA~\citep{li2020hero} and ActivityNet-QA~\citep{yu2019activitynet} are domain-specific and do not adequately capture the complexity of long-term video understanding. EgoSchema~\citep{mangalam2024egoschema} and MovieQA~\citep{tapaswi2016movieqa} provide insights into narrative and thematic understanding but are constrained by shorter video durations and limited granularity. 
While LongVideoBench~\citep{wu2024longvideobench}, MovieChat~\citep{song2023moviechat}, MoVQA~\citep{zhang2023movqa}, and Video-MME~\citep{fu2024video} utilize longer videos to test models, their average duration is still limited to around 10 minutes. In contrast, {\name} features significantly longer video segments averaging 4101 seconds, pushing the boundaries of long-term video understanding with comprehensive tasks and detailed annotations. A detailed comparison between LVBench and existing video benchmarks is presented in Table~\ref{table:dataset_compare}.

\section{{\name}}

In this chapter, we primarily discuss the construction of the original dataset for {\name} and the generation and optimization of the question-answer pairs.

\subsection{Dataset Collection}

We define long videos as those having a minimum duration of 30 minutes and containing rich, dynamic visual information with multiple events and scene transitions. To construct a comprehensive benchmark for long video understanding, we curated a diverse collection of videos from YouTube through the following systematic process.

\vpara{Video Collection.}
Starting with YouTube's search and recommendation algorithms, we gathered an initial pool of 500 videos using carefully selected keywords across different domains. Our search terms covered six primary categories: Sports Competitions (e.g., basketball matches and Olympic games), Documentary Films (e.g., nature documentaries and historical records), Event Records (e.g., ceremonies and festivals), Lifestyle and Daily Activities (e.g., cooking tutorials and travel vlogs), TV Shows and Drama Series (e.g., sitcoms and reality shows), and Cartoon Videos (e.g., animated movies and cartoon series).

\vpara{Quality Filtering.} Our annotators carefully screened these videos using five well-defined criteria:

\begin{itemize}
\item \textbf{Clear Protagonist Presence}: Videos must feature identifiable main characters (human or virtual) who appear consistently and interact meaningfully with their environment. For first-person perspective videos, the narrator's actions and decisions should be clearly observable.

\item \textbf{Structural Coherence}: Videos should maintain a well-defined narrative structure with clear beginning, development, and conclusion phases. We define ``coherent logical flow" as the presence of causally connected events that progress naturally through time.

\item \textbf{Event Density}: Videos must contain multiple distinguishable events (at least one significant event every 5 minutes on average) arranged in chronological order. We define events as meaningful actions, interactions, or state changes that contribute to the video's narrative.

\item \textbf{Visual Clarity}: The visual content should be of high quality (minimum 720p resolution) with stable camera work and clear scene compositions. We exclude videos with excessive rapid cuts, shaky footage, or poor lighting conditions.

\item \textbf{Modality Independence}: While audio may enhance understanding, all critical information should be conveyed through visual channels to ensure fair evaluation of vision-language models.
\end{itemize}

After applying these filtering criteria, we retained 103 high-quality videos totaling 117 hours of content. 

\subsection{Task Definition and Question Types}
To systematically evaluate models' capabilities in understanding long videos, we define a comprehensive taxonomy of six core skills essential for video comprehension. Each skill category is carefully designed to test specific aspects of temporal and semantic understanding while allowing for compositional combinations that assess more complex reasoning abilities.

\vpara{Temporal Grounding (TG).} These questions assess the model's ability to locate and understand events within the video's timeline, such as identifying specific moments (\textit{``What happened at 29:30?''}), determining event durations, and recognizing event sequences. This capability is fundamental for understanding the temporal dynamics in long videos.

\vpara{Summarization (Sum).} Tests the ability to synthesize and abstract information across the entire video content. Questions in this category require models to generate concise overviews, identify key developments, and understand the video's central themes, demonstrating a comprehensive understanding of long-form content.

\vpara{Reasoning (Rea).} Evaluates higher-order cognitive skills including causal understanding (\textit{``Why did the experiment fail?''}), emotional comprehension, intention interpretation, and future prediction. This category specifically examines a model's ability to make complex inferences beyond surface-level observations.

\vpara{Entity Recognition (ER).} Focuses on identifying and tracking key entities (people, objects, places) throughout the video. This includes recognizing distinct entities, understanding their relationships, tracking their actions, and connecting them to relevant events - skills crucial for following long-form narratives.

\vpara{Event Understanding (EU).} Tests comprehension of high-level semantic concepts by requiring models to classify video types, detect significant events and understand scene transitions. This capability is essential for grasping the overall structure and flow of long videos.

\vpara{Key Information Retrieval (KIR).} Evaluates attention to specific details by requiring extraction and comprehension of visual text, numerical data, and other precise information presented in the video, such as \textit{``What revenue growth did the firm report at the conference?''}

\begin{figure*}[htbp]
  \centering
    \includegraphics[width=\linewidth]{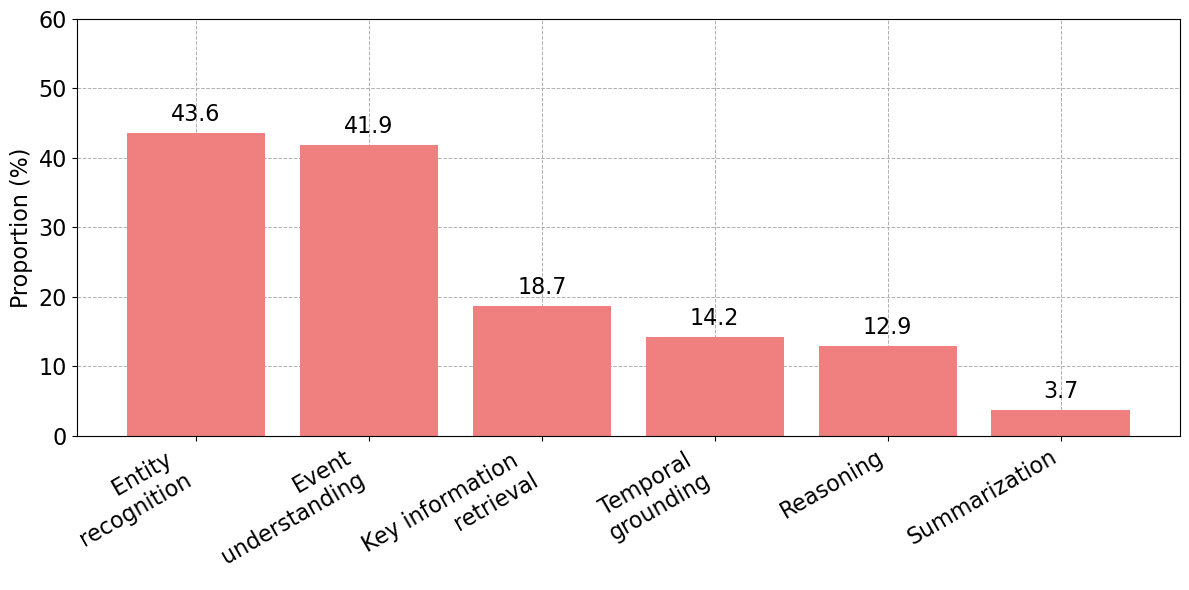}
    \vspace{-8mm}
    \caption{The proportion of different core capabilities.}
    \label{fig:category}
\end{figure*}

Figure~\ref{fig:category} provides the distribution of each question type. As shown in the figure, Entity Recognition (ER) and Event Understanding (EU) account for a larger proportion of the questions. This imbalance naturally arises from the inherent structure of video content, where almost all questions inevitably involve identifying or tracking entities and understanding events or actions. Consider the question \textit{``Why did Player number 4 in white push down Player number 17 in purple during the match?"}. While this question primarily tests reasoning capabilities (understanding the motivation behind an action), it inherently requires entity recognition (identifying and distinguishing the two players by their numbers and jersey colors) and event understanding (recognizing the pushing action within the context of the match).


\subsection{Question-Answer Pair Generation}
The annotation of long videos presents unique challenges compared to short-form content, requiring careful consideration of temporal dynamics and comprehensive coverage. We developed a systematic annotation protocol consisting of three key stages to ensure high-quality, diverse, and challenging question-answer pairs.

\vpara{Video Analysis.}
In Stage 1 (Video Analysis), annotators first watch each video in its entirety. During this initial viewing, they mark significant events, scene transitions, and key entities, while also identifying potential temporal dependencies and causal relationships that could form the basis for meaningful questions.

\vpara{Question Generation.}
Stage 2 (Question Generation) focuses on creating questions that thoroughly probe the video content. The number of questions scales with video duration, averaging 24 questions per hour. Questions are strategically distributed across the entire video duration to ensure comprehensive temporal coverage. Each video must include questions testing all six core capabilities. In constructing these questions, we emphasize specificity to ensure unique and unambiguous references to video content. For example, when questioning an argument scene, we prefer precise formulations like \textit{``When did A and B start arguing?"} or \textit{``How did the person in red's expression change during the hallway argument?"} over vague alternatives such as \textit{``Why did they start arguing?"} or \textit{``Who are the people arguing?"}. This specificity principle ensures that questions have clear, uniquely identifiable answers within the video context.

\vpara{Answer Construction.}
Stage 3 (Answer Construction) focuses on both creating answer choices and annotating temporal information needed for answering each question. For each question, annotators create four answer choices - one correct answer and three distractors. 
The correct answer precisely addresses the question, while distractors are carefully designed to share similar format and length with the correct answer. These distractors incorporate plausible but incorrect information from the video and test different aspects of understanding.

Importantly, during this stage, annotators also manually identify and mark what we term the \textbf{``clue duration"} - the minimum video segment needed to accurately answer each question. This temporal annotation involves carefully reviewing the video content to determine the exact start and end timestamps of the critical segments containing the information required for answering the question. For instance, a question about causal relationships might require a longer clue duration spanning multiple related events, while a question about a specific visual detail might have a shorter clue duration focused on a single scene. 

\begin{table*}[htbp]
    \caption{{\name} evaluation results regarding each core long video understanding capability. 
    The highest score is highlighted with green, and the second highest is highlighted with purple. 
    All the numbers are presented in \% and the full score is 100\%.}
    \centering
    \renewcommand{\arraystretch}{1.15}
    \setlength{\tabcolsep}{2mm}
    \label{tab:lvbench-results}
    \begin{tabular}{l|c|cccccccc}
        \toprule
        \textbf{Model} &  \textbf{LLM} & \textbf{ER} &  \textbf{EU} &  \textbf{KIR} &  \textbf{TG} &  \textbf{Rea} &  \textbf{Sum} &  \textbf{Overall} \\
        \midrule
        \multicolumn{5}{c}{\it{Non-Native Long Video Support Models}} \\
        \hline
        TimeChat~\citep{ren2023timechat} & LLaMA2-7B & 21.9 & 21.7 & 25.9 & 22.7 & 25.0 & 24.1 & 22.3 \\
        Video-ChatGPT~\citep{maaz2023video} & Vicuna-1.5-13B & 22.9 & 22.6	& 22.7 & 25.5 & 23.4 & 24.1	& 23.1 \\
        PLLaVA~\citep{xu2024pllava} & Yi-34B & 25.0 & 24.9 & 26.2 & 21.4 & 30.0 & 25.9 & 26.1 \\
        LLaVA-OneVision~\citep{li2024llavaone} & LLaMA3-70B & 25.0 & 26.9	& 29.2 & 30.9&	25.4 & 31.0 & 26.9 \\
        CogVLM2-Video~\citep{hong2024cogvlm2} & LLaMA3-8B & 28.3 & 26.9	&31.0 & 25.1 & 25.5	& 38.9	& 28.1\\
        LLaVA-NeXT~\citep{zhang2024llavanextvideo} & Yi-34B & 30.1 & 31.2 & 34.1 & 31.4 & 35.0 & 27.6 & 32.2 \\
        InternVL2-40B~\citep{chen2024far} & Nous-Hermes-2-Yi-34B & 37.4 & 39.7 & 43.4 & 31.4 & 42.5 & 41.4 & 39.6\\
        mPLUG-Owl3~\citep{ye2024mplug} & Qwen2-7B & 46.0 & 41.6& 42.4& 41.1 & 47.5 & 40.4 & 43.5\\
        GLM-4.1V-Thinking~\citep{hong2025glm} & GLM-4-9B& 42.8 & 43.5 & 50.0 &  39.1 & 46.0 & 25.9 & 44.3 \\
        VideoLLaMA3-7B~\citep{zhang2025videollama} & Qwen2.5-7B & 45.8 & 42.4 & 47.8 & 35.9 & 45.8 & 36.2 & 45.3\\
        GLM4V-Plus~\citep{hong2024cogvlm2} & GLM-4& 46.2 & 47.8 & 54.1 &  42.7 & 46.5 & 37.9 & 48.7 \\
        GPT-4o-20241120~\citep{gpt4o} & GPT-4o & 48.9 & 49.5 & 48.1 & 40.9 & 50.3 & 50.0 & 48.9 \\
        GPT-4.1~\citep{gpt4.1} & GPT-4.1 & - & - & - & - & - & - & 60.1 \\
        \hline
        \multicolumn{5}{c}{\it{Native Long Video Support Models}} \\
        \hline
        MovieChat~\citep{song2023moviechat} & Vicuna-7B & 21.3 & 23.1 & 25.9 & 22.3 & 24.0 & 17.2 & 22.5 \\
        LLaMA-VID~\citep{li2023llama} & Vicuna-13B & 25.4 & 21.7 & 23.4 & 26.4 & 26.5 & 17.2 & 23.9 \\
        LWM~\citep{liu2024world} & LLaMA2-7B& 24.7 & 24.8 & 26.5 & 28.6 & 30.5 & 22.4 & 25.5 \\
        Gemini-1.5-Pro~\citep{reid2024gemini} & Gemini 1.5 Pro & 32.1 & 30.9 & 39.3 & 31.8 & 27.0 & 32.8 & 33.1 \\
        Kangaroo~\citep{liu2024kangaroo} & LLaMA3-8B & 38.6 & 37.9 & 29.6 & 35.0 & 41.3 & 36.2 & 38.3 \\
        Qwen2-VL-72B~\citep{wang2024qwen2} & Qwen2-72B& 38.0 & 41.1 & 38.3 & 41.4 & 46.5 & 46.6 & 41.3\\
        Qwen2.5-VL-72B~\citep{bai2025qwen2.5} & Qwen2.5-72B & 44.2 & 40.9 & 55.6 & 37.7 & 45.2 & 34.5 & 44.0 \\
        Gemini-2.0-Flash~\citep{comanici2025gemini} & Gemini-2.0-Flash & 47.4 & 48.5 & 56.8 & 39.3 & 44.4 & 41.4 & 48.6 \\
        AdaReTaKe~\citep{wang2025adaretake} & Qwen2.5-72B & 53.0 & 50.7 & 62.2 & 45.5 & 54.7 & 37.9 & 53.3 \\
        Gemini-2.5-Flash~\citep{comanici2025gemini} & Gemini-2.5-Flash& 55.2 & 55.5 & 63.8 & 52.7 & 55.5 & 44.8 & 56.7 \\
        MR.Video~\citep{pang2025mr} & Gemini-2.0-Flash & 59.8 & 57.4 & 71.4 & 58.8 & 57.7 & 50.0 & 60.8 \\
        Seed1.5-VL~\citep{guo2025seed1} & Seed1.5 &64.3 & 64.0 & 64.7 & 52.3 & 65.0 & 51.7 & 64.0\\
        Seed1.5-VL-Thinking~\citep{guo2025seed1} & Seed1.5 & 65.4 & 63.4 & 68.0 & 53.6 & 63.7 & 46.6 & 64.6\\
        Gemini-2.5-Pro~\citep{comanici2025gemini} & Gemini-2.5-Pro & 64.5 & 67.5 & 72.8 & 65.9 & 66.5 & 58.6 & 67.4 \\
        \bottomrule
    \end{tabular}
\end{table*}

\subsection{Data Quality Control}
During the annotation process, we observed that annotators had a tendency to label most questions as temporal grounding, i.e., specifying a time range to limit the referent of the question. This practice may inadvertently reduce the difficulty of the questions and unfairly disadvantage video-understanding models that lack the ability to perceive the temporal dimension. To address this issue, we instructed annotators to minimize the number of temporal questions while still ensuring the uniqueness of the referents, effectively converting temporal grounding questions into other question types. 

Upon constructing all the questions, we discovered that for certain questions, a language model could generate answers without any visual input. As highlighted in MMstar~\citep{chen2024we}, many multimodal benchmarks can be effectively solved using pure text input alone. To mitigate this issue, we employed a straightforward yet effective approach. We utilized two powerful large language models, GLM-4~\citep{du2021glm} and GPT-4~\citep{achiam2023gpt}, to independently generate answers for all the questions. In cases where the outputs from both models were identical and matched the ground truth answer, we removed that particular data sample from the dataset. This filtering process successfully eliminated the majority of questions that did not rely on video content for answering. Following this filtering step, we obtained a refined set of 1,549 question-answer pairs.

\begin{figure*}[htbp]
  \centering
    \includegraphics[width=0.95\linewidth]{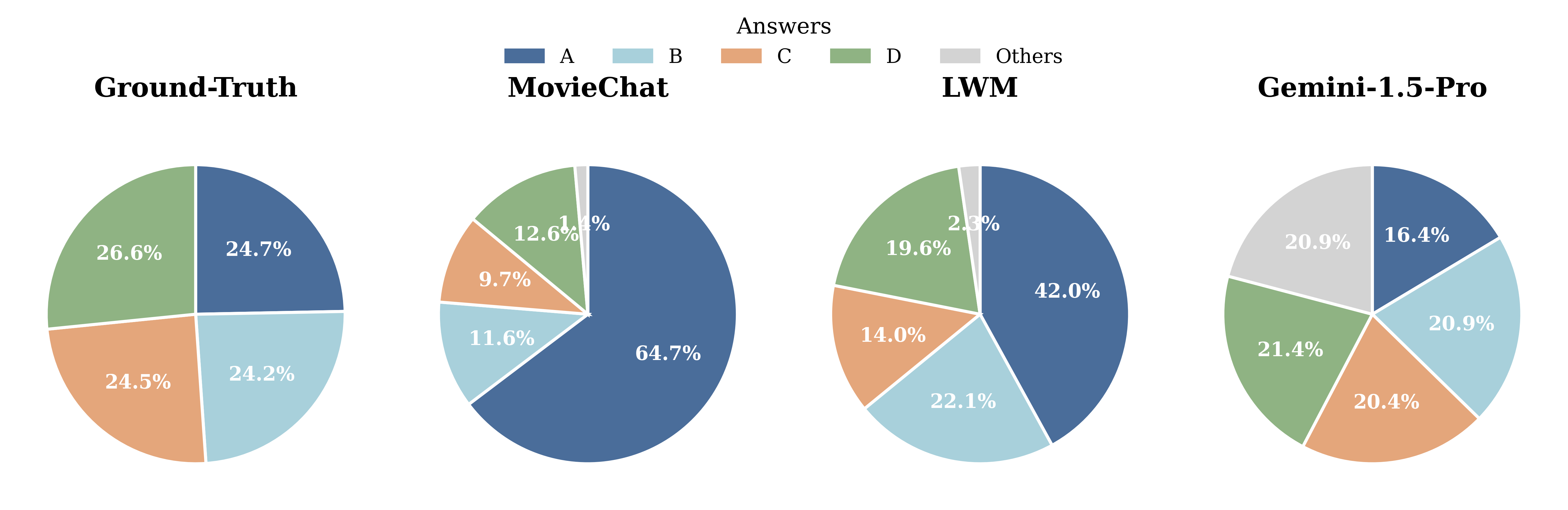}
    \vspace{-5mm}
    \caption{Failure analysis of native long-video models via answer distribution.}
    \label{fig:distribution}
\end{figure*}

\section{Experiments}

In this chapter, we report the experimental results of different video understanding models on LVBench and also compare them with human performance.

\begin{table*}[htbp]
    \caption{{\name} evaluation results across different video categories.}
    \centering
    \renewcommand{\arraystretch}{1.15}
    \setlength{\tabcolsep}{2mm}
    \label{tab:lvbench-category}
    \begin{tabular}{l|ccccccc}
        \toprule
        \textbf{Model} &  \textbf{Sports} & \textbf{Documentary} &  \textbf{Event Record} &  \textbf{Lifestyle} &  \textbf{TV Show} &  \textbf{Cartoon} &  \textbf{Overall} \\
        \midrule
        Human & 96.3 & 89.8 & 87.4 & 98.4 & 97.2 & 95.8 & 94.4\\
        Seed1.5-VL & 63.3 & 67.5 & 60.3 & 66.9 & 60.4 & 65.0 & 64.0 \\
        Gemini-2.5-Pro &  71.3 & 54.3 & 72.1 & 68.5 & 69.2 & 66.1 & 67.4\\
        \bottomrule
    \end{tabular}
\end{table*}

\subsection{Settings}
We evaluated the performance of 13 models that support multi images or short video input: TimeChat~\cite{ren2023timechat}, Video-ChatGPT~\citep{maaz2023video}, PLLaVA~\cite{xu2024pllava}, LLaVA-OneVision~\citep{li2024llavaone}, CogVLM2-Video~\citep{hong2024cogvlm2}, LLaVA-NeXT~\citep{zhang2024llavanextvideo},   InternVL2-40B~\citep{chen2024far}, mPLUG-Owl3~\citep{ye2024mplug}, GLM-4.1V-Thinking~\citep{hong2025glm}, VideoLLaMA3-7B~\citep{zhang2025videollama}, GLM4V-Plus~\citep{hong2024cogvlm2}, GPT-4o~\citep{gpt4o} and GPT-4.1~\citep{gpt4.1}. To adapt these models for long video inputs, we sample a fixed number of frames from the original video, such as 32 or 96 frames, to maintain consistency with the model's training sequence length. Additionally, we assessed 13 models that natively support long videos: LLaMA-VID~\citep{li2023llama}, MovieChat~\citep{song2023moviechat}, LWM~\citep{liu2024world}, Gemini 1.5 Pro~\citep{reid2024gemini}, Kangaroo~\citep{liu2024kangaroo}, Qwen2-VL-72B~\citep{wang2024qwen2}, Qwen2.5-VL-72B~\citep{bai2025qwen2.5}, Gemini-2.0-Flash~\citep{comanici2025gemini}, AdaReTaKe~\citep{wang2025adaretake}, Gemini-2.5-Flash~\citep{comanici2025gemini}, MR.Video~\citep{pang2025mr}, Seed1.5-VL~\citep{guo2025seed1} and Gemini-2.5-Pro~\citep{comanici2025gemini}. We processed the videos at a rate of 1 frame per second and fed them into the models, only performing downsampling when the video's length exceeded the model's maximum processing capability. It is worth noting that although Gemini 1.5 Pro can handle videos up to 10 hours long, its publicly available interface is limited to processing videos of up to 1 hour in length.
For each question, we provided the following prompt as input to the models:
\begin{displayquote}
\textit{{Question} (A) {Option1} (B) {Option2} (C) {Option3} (D) {Option4}.
Please select the best answer from the options above and directly provide the letter representing your choice without giving any explanation.}
\end{displayquote}
After obtaining the model responses, we first attempted to extract the answers using regular expression matching. For questions where the matching process was unsuccessful, we employed a LLM to extract the answers from the responses.

\subsection{Performance across Core Capabilities}
To comprehensively evaluate the performance of various long video understanding models across core capabilities, we conducted extensive experiments on the {\name} dataset, testing multiple representative models, including both non-native and native long video support models. The experimental results are presented in Table \ref{tab:lvbench-results}. 

Overall, Gemini-2.5-Pro demonstrated state-of-the-art (SOTA) performance, achieving a top score of 67.4 by outperforming all other models in 5 out of 6 tasks (EU, KIR, TG, Rea, and Sum). Seed1.5-VL-Thinking secured the second position with a score of 64.6. Our analysis reveals two key findings. First, a significant performance gap persists between proprietary and open-source models. The top-performing open-source models, VideoLLaMA3-7B (non-native long-video) and Qwen2.5-VL-72B (native long-video), scored 45.3 and 44.0, respectively, lagging behind Gemini-2.5-Pro by over 20\% in both individual tasks and the overall score. Second, we observed that some models without native long-video support still achieved competitive results against their native counterparts, though their performance was not comparable to SOTA models. Conversely, several models designed for native long-video processing, including MovieChat, LLaMA-VID, LWM, and Gemini-1.5-Pro, exhibited unexpectedly low scores.


\subsection{Analysis of Failure Modes}
To understand why some native long video support models perform poorly on LVBench, we evaluated the distribution of answers generated by different models on {\name} and observed that existing long video understanding models struggle with precisely following instructions.  For example, despite explicitly constraining the output in the prompt to be one of four provided answer choices, Gemini-1.5-Pro generated responses outside of the specified options 20.9\% of the time, such as \textit{"None of the above options are correct"} or \textit{"I cannot answer this question"}. This occurred even though manual validation confirmed that the questions were indeed answerable from the given choices. MovieChat and LWM exhibited a strong bias towards selecting option A, regardless of the question. 

We hypothesize that although these models have the ability to process long videos, they may lack instruction data for such long videos, which leads to them being unable to effectively follow instructions, in turn affecting their performance.

\subsection{Performance across Video Categories}
We conducted a comprehensive evaluation across various video categories. We selected two leading models, Seed1.5-VL and Gemini-2.5-Pro, for detailed analysis and compared their results with human performance.

As shown in Table~\ref{tab:lvbench-category}, humans achieve a very high accuracy of 94.4\% on average, setting a strong benchmark across all categories. In contrast, the overall performance of Gemini-2.5-Pro and Seed1.5-VL was considerably lower, at 67.4\% and 64.0\%, respectively. This highlights a significant performance gap between current multimodal models and humans in long-video understanding, indicating substantial room for improvement.

Further analysis of the results by category reveals distinct model strengths. Gemini-2.5-Pro performed best on Event Record videos, reaching an accuracy of 72.1\%, but struggled with Documentary videos, where its score dropped to 54.3\%. Conversely, Seed1.5-VL demonstrated more consistent performance across categories, with all scores falling within a narrow range between 60.3\% (Event Records) and 67.5\% (Documentaries).

\subsection{Ablation on Frame Density}

\begin{figure}[htbp]
  \centering
    \includegraphics[width=\linewidth]{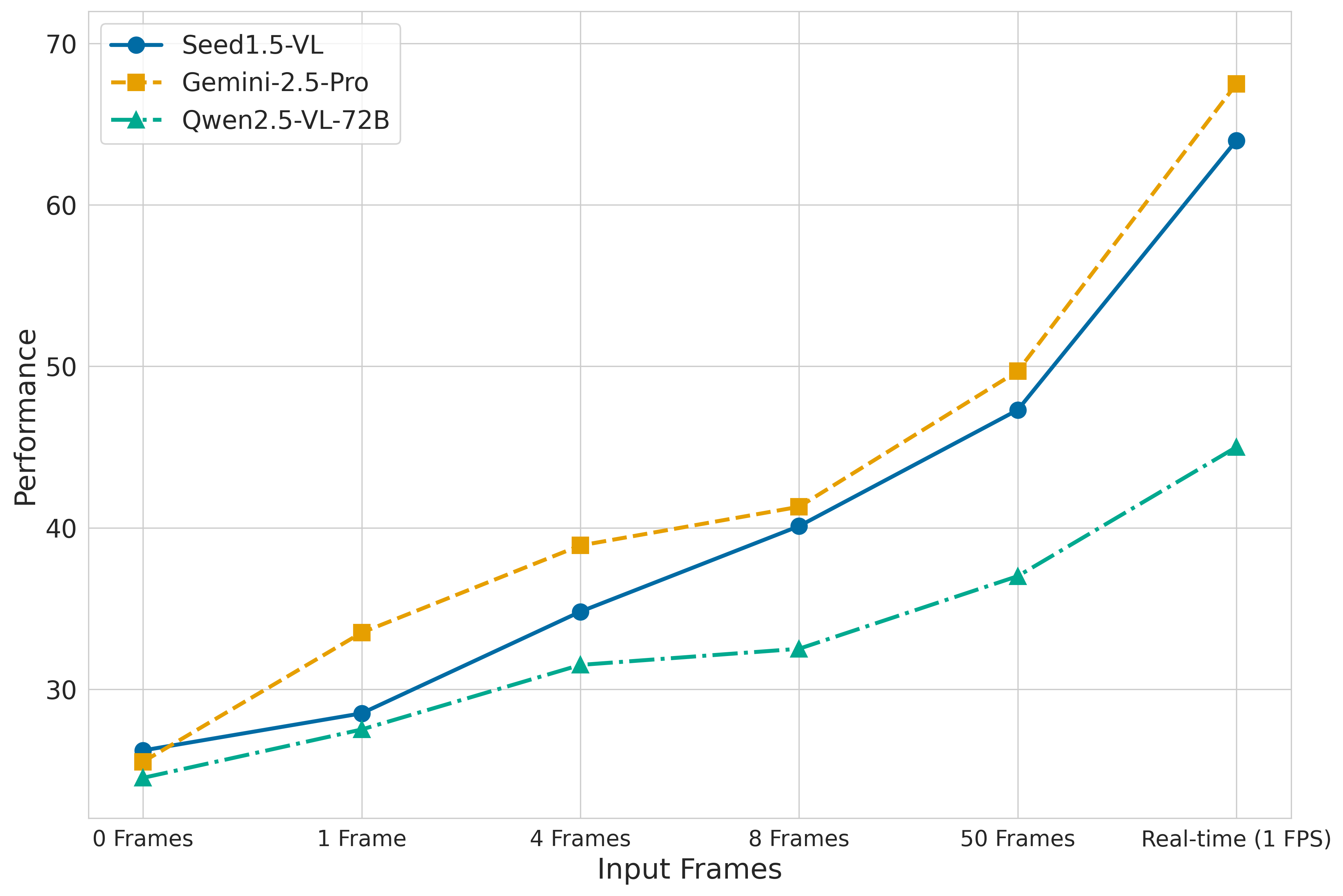}
    \vspace{-2mm}
    \caption{Impact of Frame Count on Model Performance.}
    \label{fig:frame}
    \vspace{-2mm}
\end{figure}

To understand the relationship between temporal information density and model performance, we conduct an ablation study on the number of input frames. We evaluate three state-of-the-art models—Seed1.5-VL, Gemini-2.5-Pro, and Qwen2.5-VL-72B—with varying frame inputs: 0, 1, 4, 8, 50, and a dense sampling of 1 frame per second (FPS). The results are presented in Figure ~\ref{fig:frame}.

As depicted in the figure, all models exhibit a clear trend of improved performance with an increased number of input frames. A modest performance gain is observed as the frame count increases from 1 to 8. This suggests that with sparse frames, models can grasp the general context or primary event of the video but may fail to capture the transient, crucial visual cues necessary for accurate reasoning. The most significant finding is the substantial performance leap from 50 frames to the 1 FPS setting. This large gap underscores the necessity of dense visual input for resolving the complex, long-range temporal dependencies inherent in our benchmark tasks.

Furthermore, the ``0 Frames" condition serves as a critical control experiment to validate the integrity of our benchmark's question-answer pairs. In this visuals-agnostic setting, models must attempt to answer based solely on linguistic cues from the question and options. The results show that all models perform near the level of random guessing. This outcome effectively demonstrates that success on our benchmark is contingent on visual understanding rather than exploitable language biases. It also validates the efficacy of our LLM-based data filtering protocol, which successfully purges questions that do not strictly require visual grounding.

\section{Discussion}
\vpara{Conclusion.}In this paper, we introduced {\name}, a benchmark designed to advance long video understanding. {\name} comprises a diverse collection of lengthy videos and a meticulously annotated question-answer dataset, presenting a robust evaluation framework for assessing multimodal models on complex video understanding tasks. Our experiments revealed that while state-of-the-art models have made strides in short video understanding, their performance on long videos falls short of human-level accuracy. 
By providing a challenging benchmark, we hope to stimulate the development of advanced models capable of tackling the complexities of extended video comprehension.

\vpara{Limitations.} A limitation of our benchmark is the exclusion of audio data. While audio can provide valuable context, we did not include it because most current models lack effective audio processing capabilities. Future work will aim to incorporate audio information to enhance the evaluation framework for multimodal understanding.

\vpara{License.} Our dataset is under the CC-BY-NC-SA-4.0 license. The dataset relies on YouTube as an external resource for the video content. We only provide download links rather than source videos. While there is no guarantee that the original YouTube videos will remain constant over time, we have taken measures to mitigate this issue by creating and storing copies of the videos. These copies can be provided to dataset consumers, subject to obtaining permission from the original video creators.

\subsubsection*{Acknowledgments}
This work was supported by the National Natural Science Foundation of China (NSFC) under Grants 62425601 and 62495063, and by the New Cornerstone Science Foundation through the XPLORER PRIZE.

{
    \small
    \bibliographystyle{ieeenat_fullname}
    \bibliography{main}
}
\clearpage

\appendix
\section{Datasheet} \label{ap:datasheet}
\subsection{Motivation}

\begin{itemize}

\item \textit{\textbf{For what purpose was the dataset created?} Was there a specific task in mind? Was there a specific gap that needed to be filled? Please provide a description.}\\
The dataset was created to evaluate and optimize long video understanding models. To the best of our knowledge, this is the first benchmark capable of assessing models' performance on extra-long videos.

\item \textit{\textbf{Who created the dataset (e.g., which team, research group) and on behalf of which entity (e.g., company, institution, organization)?}}\\
\textit{Anonymous for review}

\item \textit{\textbf{Who funded the creation of the dataset?} If there is an associated grant, please provide the name of the grantor and the grant name and number.}\\
\textit{Anonymous for review}

\item \textit{\textbf{Any other comments?}}\\
No.

\end{itemize}

\subsection{Composition}

\begin{itemize}

\item \textit{\textbf{What do the instances that comprise the dataset
    represent (e.g., documents, photos, people, countries)?} Are there
  multiple types of instances (e.g., movies, users, and ratings;
  people and interactions between them; nodes and edges)? Please
  provide a description.}\\
The instances comprising the dataset are YouTube videos, which can be categorized into six different types: Sports, Documentary, Event Record, Lifestyle, TV Show, and Cartoon.

\item \textit{\textbf{How many instances are there in total (of each type, if appropriate)?}}\\
In total, there are 103 videos and 1,549 question-answer pairs in the dataset.

\item \textit{\textbf{Does the dataset contain all possible instances or is it
    a sample (not necessarily random) of instances from a larger set?}
  If the dataset is a sample, then what is the larger set? Is the
  sample representative of the larger set (e.g., geographic coverage)?
  If so, please describe how this representativeness was
  validated/verified. If it is not representative of the larger set,
  please describe why not (e.g., to cover a more diverse range of
  instances, because instances were withheld or unavailable).}\\
The dataset contains a sample of instances from a larger set of YouTube videos and their corresponding question-answer pairs. The multi-stage filtering process was employed to select high-quality question-answer pairs for inclusion in the dataset.

\item \textit{\textbf{What data does each instance consist of?} ``Raw'' data
  (e.g., unprocessed text or images) or features? In either case,
  please provide a description.}\\
Each instance in the dataset consists of a single video and its corresponding supplementary information. All the additional information is stored in the ``video\_info.meta.jsonl" file, where:
\begin{itemize}
    \item All question-answer pairs associated with the video are located under the ``qa" key.
    \item The video category is specified under the ``type" key.
    \item Details such as video duration, resolution, and other metadata are contained within the ``video\_info" key.
\end{itemize}

\item \textit{\textbf{Is there a label or target associated with each
    instance?} If so, please provide a description.}\\
Yes, each video instance in the dataset is associated with a set of corresponding question-answer pairs, which serve as the labels or targets for that particular instance.

\item \textit{\textbf{Is any information missing from individual instances?}
  If so, please provide a description, explaining why this information
  is missing (e.g., because it was unavailable). This does not include
  intentionally removed information, but might include, e.g., redacted
  text.}\\
No.

\item \textit{\textbf{Are relationships between individual instances made
    explicit (e.g., users' movie ratings, social network links)?} If
  so, please describe how these relationships are made explicit.}\\
N/A.

\item \textit{\textbf{Are there recommended data splits (e.g., training, development/validation, testing)?} If so, please provide a description of these splits, explaining the rationale behind them.}\\
No, there are no recommended data splits for this dataset. The entire dataset is intended to be used for evaluating and testing the performance of long video understanding models.

\item \textit{\textbf{Are there any errors, sources of noise, or redundancies
    in the dataset?} If so, please provide a description.}\\
Yes, there is a small probability of bias introduced by the annotators during the labeling process.

\item \textit{\textbf{Is the dataset self-contained, or does it link to or
    otherwise rely on external resources (e.g., websites, tweets,
    other datasets)?} If it links to or relies on external resources,
    a) are there guarantees that they will exist, and remain constant,
    over time; b) are there official archival versions of the complete
    dataset (i.e., including the external resources as they existed at
    the time the dataset was created); c) are there any restrictions
    (e.g., licenses, fees) associated with any of the external
    resources that might apply to a dataset consumer? Please provide
    descriptions of all external resources and any restrictions
    associated with them, as well as links or other access points, as
    appropriate.}\\
The dataset relies on YouTube as an external resource for the video content. While there is no guarantee that the original YouTube videos will remain constant over time, we have taken measures to mitigate this issue by creating and storing copies of the videos. These copies can be provided to dataset consumers, subject to obtaining permission from the original video creators.

\item \textit{\textbf{Does the dataset contain data that might be considered
    confidential (e.g., data that is protected by legal privilege or
    by doctor--patient confidentiality, data that includes the content
    of individuals' non-public communications)?} If so, please provide
    a description.}\\
No.

\item \textit{\textbf{Does the dataset contain data that, if viewed directly,
    might be offensive, insulting, threatening, or might otherwise
    cause anxiety?} If so, please describe why.}\\
No.

\item \textit{\textbf{Any other comments?}}\\
No.

\end{itemize}

\subsection{Collection Process}

\begin{itemize}

\item \textit{\textbf{How was the data associated with each instance
    acquired?} Was the data directly observable (e.g., raw text, movie
  ratings), reported by subjects (e.g., survey responses), or
  indirectly inferred/derived from other data (e.g., part-of-speech
  tags, model-based guesses for age or language)? If the data was reported
  by subjects or indirectly inferred/derived from other data, was the
  data validated/verified? If so, please describe how.}\\
The data associated with each instance in the dataset was manually annotated by hired annotators.

\item \textit{\textbf{What mechanisms or procedures were used to collect the
    data (e.g., hardware apparatus(es) or sensor(s), manual human
    curation, software program(s), software API(s))?} How were these
    mechanisms or procedures validated?}\\
The data collection process involved the use of dedicated servers to download the YouTube videos, followed by the annotation of the videos using an annotation platform.

\item \textit{\textbf{If the dataset is a sample from a larger set, what was
    the sampling strategy (e.g., deterministic, probabilistic with
    specific sampling probabilities)?}}\\
See the third question of the previous section on ``Composition".

\item \textit{\textbf{Who was involved in the data collection process (e.g.,
    students, crowdworkers, contractors) and how were they compensated
    (e.g., how much were crowdworkers paid)?}}\\
The data annotation process was carried out by professional annotators, who were compensated with an average of 30 US dollars per video annotated.

\item \textit{\textbf{Over what timeframe was the data collected?} Does this
  timeframe match the creation timeframe of the data associated with
  the instances (e.g., recent crawl of old news articles)?  If not,
  please describe the timeframe in which the data associated with the
  instances was created.}\\
\textit{Anonymous for review}

\item \textit{\textbf{Were any ethical review processes conducted (e.g., by an
    institutional review board)?} If so, please provide a description
  of these review processes, including the outcomes, as well as a link
  or other access point to any supporting documentation.}\\
No.

\item \textit{\textbf{Any other comments?}}\\
No.

\end{itemize}

\subsection{Preprocessing/cleaning/labeling}

\begin{itemize}

\item \textit{\textbf{Was any preprocessing/cleaning/labeling of the data done
    (e.g., discretization or bucketing, tokenization, part-of-speech
    tagging, SIFT feature extraction, removal of instances, processing
    of missing values)?} If so, please provide a description. If not,
  you may skip the remaining questions in this section.}\\
Yes, preprocessing and cleaning of the data was performed. We manually inspected the content of each video to ensure all videos were high-quality, content-rich, and free of harmful information.

\item \textit{\textbf{Was the ``raw'' data saved in addition to the preprocessed/cleaned/labeled data (e.g., to support unanticipated future uses)?} If so, please provide a link or other access point to the ``raw'' data.}\\
No.

\item \textit{\textbf{Is the software that was used to preprocess/clean/label the data available?} If so, please provide a link or other access point.}\\
No.

\item \textit{\textbf{Any other comments?}}\\
No. 

\end{itemize}

\subsection{Uses}

\begin{itemize}

\item \textit{\textbf{Has the dataset been used for any tasks already?} If so, please provide a description.}\\
Not beyond this paper.

\item \textit{\textbf{Is there a repository that links to any or all papers or systems that use the dataset?} If so, please provide a link or other access point.}\\
\textit{Anonymous for review}

\item \textit{\textbf{What (other) tasks could the dataset be used for?}}\\
The dataset could be used for various other long video understanding tasks such as summarization, captioning, question answering and multimodal understanding.

\item \textit{\textbf{Is there anything about the composition of the dataset or the way it was collected and preprocessed/cleaned/labeled that might impact future uses?} For example, is there anything that a dataset consumer might need to know to avoid uses that could result in unfair treatment of individuals or groups (e.g., stereotyping, quality of service issues) or other risks or harms (e.g., legal risks, financial harms)? If so, please provide a description. Is there anything a dataset consumer could do to mitigate these risks or harms?}\\
No.

\item \textit{\textbf{Are there tasks for which the dataset should not be used?} If so, please provide a description.}\\
No.

\item \textit{\textbf{Any other comments?}}\\
No.

\end{itemize}

\subsection{Distribution}

\begin{itemize}

\item \textit{\textbf{Will the dataset be distributed to third parties outside of the entity (e.g., company, institution, organization) on behalf of which the dataset was created?} If so, please provide a description.}\\
Yes, the dataset is freely and publicly available and accessible.

\item \textit{\textbf{How will the dataset be distributed (e.g., tarball on website, API, GitHub)?} Does the dataset have a digital object identifier (DOI)?}\\
\textit{Anonymous for review}

\item \textit{\textbf{When will the dataset be distributed?}}\\
\textit{Anonymous for review}

\item \textit{\textbf{Will the dataset be distributed under a copyright or other intellectual property (IP) license, and/or under applicable terms of use (ToU)?} If so, please describe this license and/or ToU, and provide a link or other access point to, or otherwise reproduce, any relevant licensing terms or ToU, as well as any fees associated with these restrictions.}\\
Our dataset is under the CC-BY-NC-SA-4.0 license.

\item \textit{\textbf{Have any third parties imposed IP-based or other restrictions on the data associated with the instances?} If so, please describe these restrictions, and provide a link or other access point to, or otherwise reproduce, any relevant licensing terms, as well as any fees associated with these restrictions.}\\
No.

\item \textit{\textbf{Do any export controls or other regulatory restrictions apply to the dataset or to individual instances?} If so, please describe these restrictions, and provide a link or other access point to, or otherwise reproduce, any supporting documentation.}\\
No.

\item \textit{\textbf{Any other comments?}}\\
No.

\end{itemize}

\subsection{Maintenance}

\begin{itemize}

\item \textit{\textbf{Who will be supporting/hosting/maintaining the dataset? }}\\
\textit{Anonymous for review}

\item \textit{\textbf{How can the owner/curator/manager of the dataset be contacted (e.g., email address)?}}\\
\textit{Anonymous for review}

\item \textit{\textbf{Is there an erratum?} If so, please provide a link or other access point.}\\
No.

\item \textit{\textbf{Will the dataset be updated (e.g., to correct labeling
    errors, add new instances, delete instances)?} If so, please
  describe how often, by whom, and how updates will be communicated to
  dataset consumers (e.g., mailing list, GitHub)?}\\
\textit{Anonymous for review}

\item \textit{\textbf{If the dataset relates to people, are there applicable
    limits on the retention of the data associated with the instances
    (e.g., were the individuals in question told that their data would
    be retained for a fixed period of time and then deleted)?} If so,
    please describe these limits and explain how they will be
    enforced.}\\
Yes, there are applicable limits on the retention of data associated with individuals in the videos. If the original creators of the videos believe their content should be removed from the dataset, we will remove it upon request.

\item \textit{\textbf{Will older versions of the dataset continue to be
    supported/hosted/maintained?} If so, please describe how. If not,
  please describe how its obsolescence will be communicated to dataset
  consumers.}\\
\textit{Anonymous for review}

\item \textit{\textbf{If others want to extend/augment/build on/contribute to
    the dataset, is there a mechanism for them to do so?} If so,
  please provide a description. Will these contributions be
  validated/verified? If so, please describe how. If not, why not? Is
  there a process for communicating/distributing these contributions
  to dataset consumers? If so, please provide a description.}\\
No, there is no mechanism for others to extend, augment, build on, or contribute to the dataset. This decision was made to ensure consistency and fairness in benchmarking and evaluation using the dataset.

\item \textit{\textbf{Any other comments?}}\\
No.

\end{itemize}

\begin{figure*}[htbp] 
    \centering
    \rotatebox{90}{%
        \includegraphics[width=\textheight]{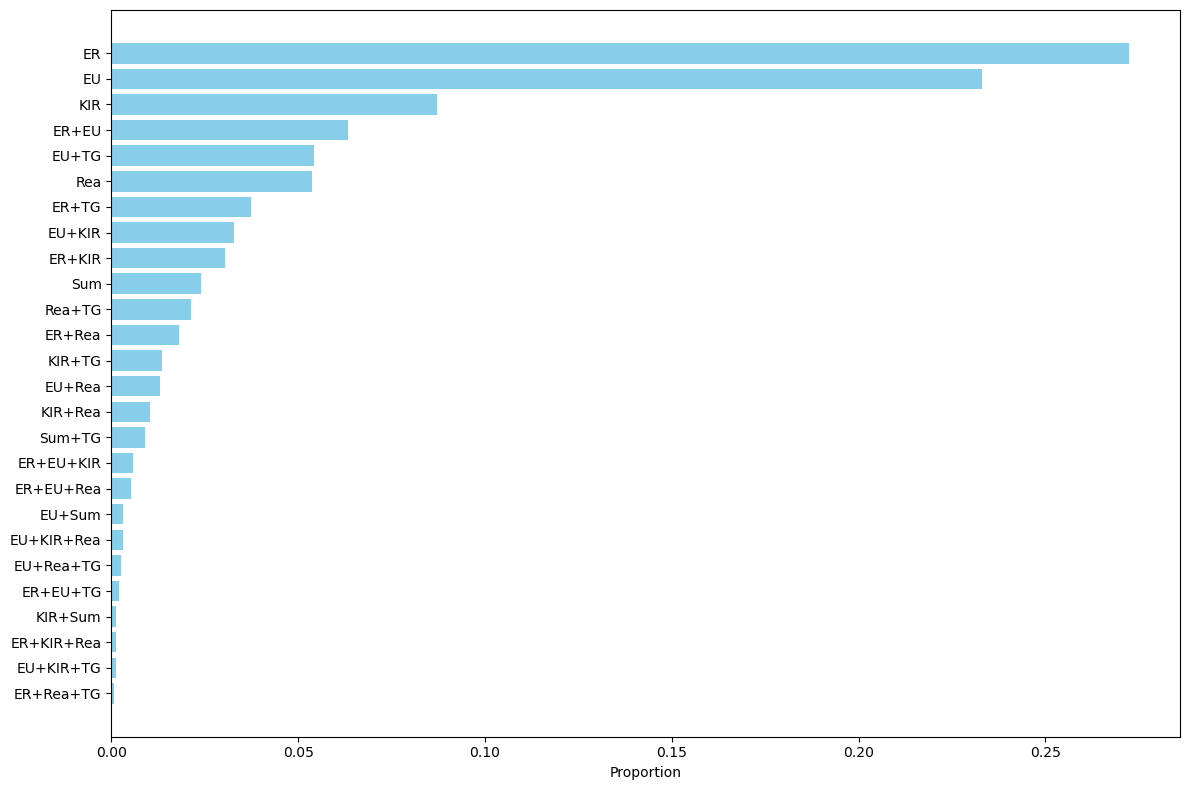} 
    }
    \caption{The proportion of core capability combinations.}
    \label{fig:coarse_category}
\end{figure*}

\section{Distribution of Core Capability Combinations}
In this section, we quantified the distribution of various combinations of core competencies within the dataset. As illustrated in the Figure~\ref{fig:coarse_category}, the six core competencies can be further combined to form 26 fine-grained question types. This flexible combinatorial approach guarantees the richness and diversity of the dataset, enabling a comprehensive evaluation of the model's performance across multiple dimensions.

\end{document}